\documentclass[runningheads]{llncs}
\usepackage{graphicx}
\usepackage{booktabs}
\usepackage{amsmath,amssymb}
\usepackage{booktabs}
\usepackage{url}
\begin{document}
\title{Predicting Research Trends From Arxiv}
%
%
\author{Steffen Eger\inst{1} 
\and
Chao Li\inst{2} \and
Florian Netzer\inst{1} \and
Iryna Gurevych\inst{1}
}
\authorrunning{S.~Eger et al.}
%
\institute{Technical University Darmstadt, Darmstadt, Germany \and
Frankfurt School of Finance and Management, Frankfurt am Main, Germany. 
}
\maketitle              
\begin{abstract}
%
We perform trend detection on two datasets of Arxiv papers, derived from its machine learning (cs.LG) and natural language processing (cs.CL) categories. Our approach is bottom-up: we first rank papers by their normalized citation counts, then group top-ranked papers into different categories based on the tasks that they pursue and the methods they use. We then analyze these resulting topics. We find that the dominating paradigm in cs.CL revolves around \emph{natural language generation} problems and those in cs.LG revolve around \emph{reinforcement learning} and \emph{adversarial principles}. By extrapolation, we predict that these topics will remain lead problems/approaches in their fields in the short- and mid-term.  
\end{abstract}
\section{Introduction}
Predicting trends in research has been a long-standing dream of scientists. Projects on popular research topics often lead to higher acceptance rates at conferences and journals, as well as funding application approvals. 
Further, knowing future research trends 
also has implications for society as a whole, 
because these trends will most likely directly affect the labor market, technological orientation and biases, consumer end products, as well as cultural metaphors and definitions of the human identity---this is even more true for fields such as artificial intelligence, as we focus on here. 
However, with the accelerating number of papers made available each year, it becomes ever more difficult to digest the incoming information and thereby manually 
identify topics that will have 
long-term scientific impact. 
We have developed an automatic system whose goal is to uncover  important research trends, and, therefore, aims at helping  researchers better plan their academic endeavors. 

Our system crawls 
papers published in the Machine Learning (cs.LG) and natural language processing (cs.CL) categories of Arxiv,\footnote{https://arxiv.org/} 
with information about how often they were cited. In this dataset, we identify promising papers using normalized citation counts and then categorize them by hand and automatically.
Using Arxiv papers for our exploration appears promising, because Arxiv is a very popular pre-print (and post-print) server for scientific publications, whose impact has, moreover, considerably increased over the last few years.\footnote{See submission statistics to Arxiv at \url{https://arxiv.org/help/stats/2018_by_area/index}.}

\section{Data \& Annotation}
\subsection{Data}
We created two datsets: One with papers from the machine learning (cs.LG) category of Arxiv and one with papers from computation and language (cs.CL). 
We focus on these two prominent subfields of artificial intelligence because they appear concurrently particularly dynamic, with drastic changes and performance improvements witnessed each year, mainly due to the impact of artificial neural network (aka deep learning) models. 
The data includes papers with their title, abstract and authors. We also harvested citation information for the papers from semanticscholar.\footnote{https://www.semanticscholar.org/} 
Our crawled papers date between June, 2017 and December, 2018. We crawled roughly 4.8k papers from cs.CL and 12.4k papers from cs.LG.
\subsection{Annotation}
One co-author of this study manually annotated the abstracts of the top-100 (according to their normalized citation count; see Section \ref{sec:zscore}) papers in cs.LG and cs.CL, respectively, for three aspects: \texttt{task}, \texttt{method}, and \texttt{goal/findings}. These aspects answer the questions \emph{what} a paper researches, and \emph{how} and \emph{why} it does so.

\begin{table}[!htb]
{\footnotesize
    \centering
    \begin{tabular}{l|ccc}
         \toprule 
         Title & \texttt{task} & \texttt{method} & \texttt{goal}  \\ \toprule
         IEST: WASSA-2018 (18, 13.0)  & Emotion Det. & Data & Difficult task \\ 
         Implicit Emotions Shared Task & \\
         \midrule
         BERT: Pre-training of Deep (20, 12.0) & Text repr.  & Transformer & Better \\ Bidirectional Transformers for & & & accuracy\\
         Language   Understanding \\ \midrule
         Deep contextualized (261, 11.2) & Text repr. & Language Model & Better \\
         word representations & & & accuracy \\ \bottomrule
    \end{tabular}
    \caption{Top-3 papers from cs.CL according to their normalized citation counts, as well as \texttt{task}, \texttt{method} and \texttt{goal} classification. In brackets: absolute and normalized citations by end of December 2018.}
    \label{table:examples}
    }
\end{table}
Table \ref{table:examples} lists the top-3 papers from cs.CL, together with their \texttt{task}, \texttt{method}, and \texttt{goal} annotations. For cs.CL, we decided upon 15 fine-grained \texttt{task} label categories, given in Table \ref{table:task_csCL}. Similarly, we decided upon 28 \texttt{method} categories, and 7 \texttt{goal} categories. For cs.LG, we used 13,15, and 13 \texttt{task}, \texttt{method}, and \texttt{goal} labels, respectively. We note that it was sometimes difficult to annotate the  abstracts for any of the three categories because the information in question may not have been available from the abstract or an abstract could not be clearly assigned to one of the agreed upon categories. We also note that we used different labeling schemes for cs.LG and cs.CL, respectively. This is because papers in cs.CL often deal with a fine-grained natural language processing (NLP) problem whereas wording in cs.LG abstracts ignored such fine-grained differences or claimed to address \emph{various} problems from, e.g., image or text fields. We also mention the general vagueness 
and arbitrariness (to a certain degree) of such classifications. Overall, it was easiest and least ambiguous to classify cs.CL papers to \texttt{tasks} and cs.LG papers to \texttt{methods}. Hence, we mostly focus on these two combinations.

\begin{table}[!htb]
    \centering
    \begin{tabular}{ll}
    \toprule
         \texttt{task}-cs.CL &  Generation, Machine Translation (MT), Text representations, Speech, \\
         & Language Modeling, Sentence Classification, Style Transfer, Reasoning,\\ & Relation extraction, Sequence Tagging, Emotion Detection, \\
         & Argument Mining, Human-Computer Interaction, Parsing, Rest\\
         \texttt{method}-cs.LG & Reinforcement Learning (RL), Other Deep Learning architect., \\
         & Representation Learning, GAN, Generation, Architecture Search,\\
         & Distillation, Analysis, Interpretability, Learning Aspects,\\ &Various, Data, Rest, Adversarial\\
         \bottomrule
    \end{tabular}
    \caption{\texttt{task} and \texttt{method} labels for cs.CL and cs.LG, respectively.}
    \label{table:task_csCL}
\end{table}

\section{Normalized citation counts}
The simplest measure for a paper's influence is the number of citations it has received. But plain citation counts may be misleading. They may vary depending on the research field and the date of the publication.
Instead, citation counts can be normalized by comparing only papers in the same research fields and adjusting citation count scores by the paper's age.  
This is the idea of the \emph{z-score} approach suggested by Newman  \cite{Newman:2009,Newman:2014}. 
This is calculated by subtracting the mean citation count of papers within a time window from the citation count of a paper and dividing by the standard deviation.  
With this method, Newman \cite{Newman:2014} does not merely identify promising papers as those that have a high number of citations because they were published earlier than other papers from the same field, but also papers with only a few citations.
To illustrate the effectiveness of the method, after 5 years, the 50 papers with the initially highest z-scores in Newman's sample received 15 times as many citations as the average paper in a randomly drawn control group \emph{that started with the same number of citations}. Thus, Newman argues 
that the z-score can indeed identify short- and mid-term research trends.\footnote{We assume here that high future citation counts are a proxy for the research trend prediction task, and ignore cases of, e.g., negative citations.} 

To find promising papers, we calculated the z-score for the papers in our datasets, using a time-window of  $\pm$10 days. 
We 
ignored papers with less than 4 citations, because we deemed such small numbers  unreliable.

\label{sec:zscore}
\section{Results}

\begin{figure}
    \centering
    \includegraphics[scale=0.7]{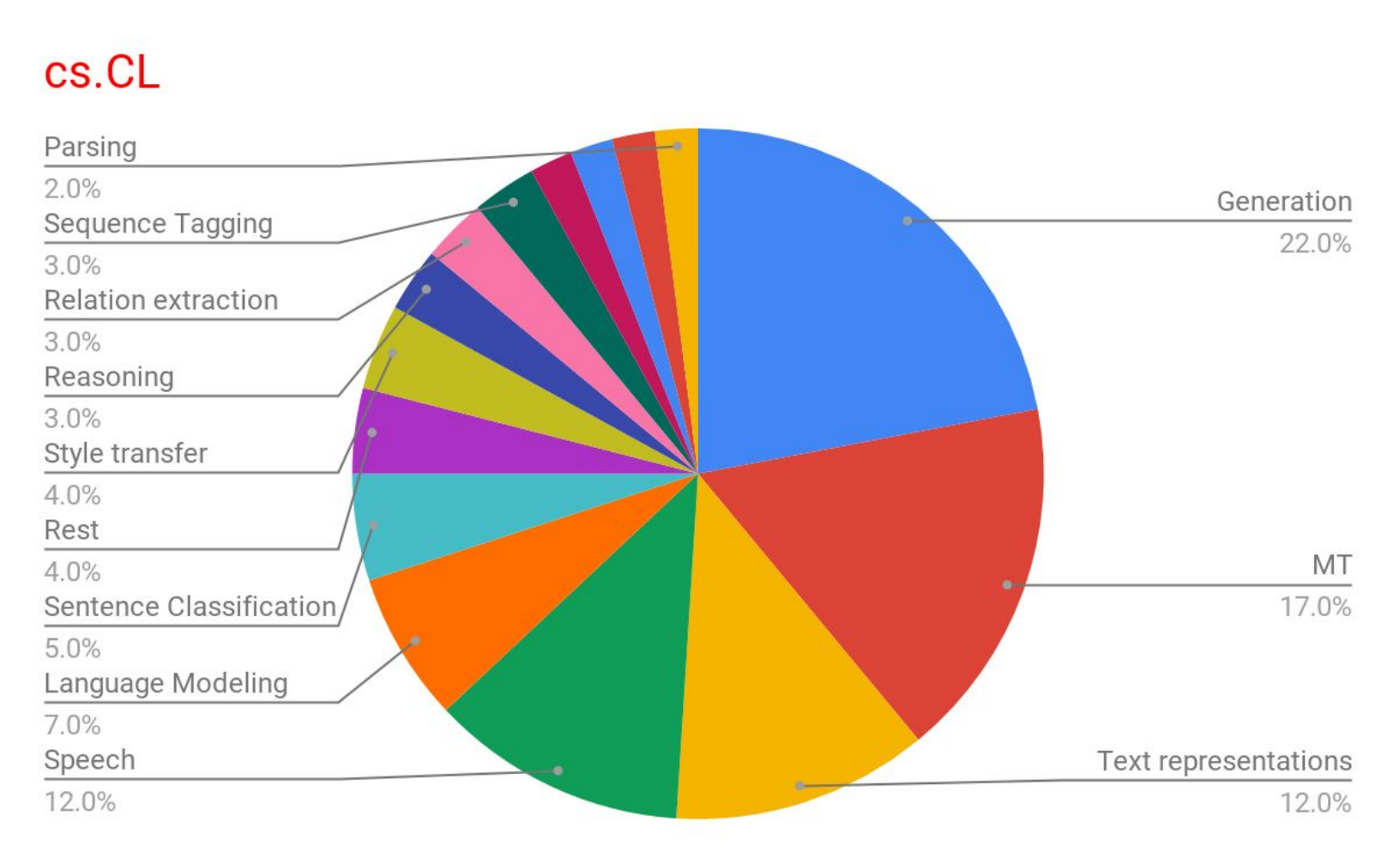}
    \caption{\texttt{task} distribution in top-100 list in cs.CL.}
    \label{figure:csCL}
\end{figure}

Figure \ref{figure:csCL} shows the distribution of \texttt{task}s in the top-100 list of cs.CL, with topic labels given in Table \ref{table:task_csCL}. There are few very interesting patterns: for example, traditional NLP tasks like sequence tagging and parsing are marginalized and make up only 5\% of the top papers, while the most prominent tasks dealt with in the top-100 papers are almost all about natural language generation: machine translation, language modeling (where the goal is to predict the next word given the previous text sequence) and generation proper (consisting of subcategories dialogue, question answering, text2SQL generation and summarization). The \texttt{task} speech mostly also deals with generation, namely, generation of a speech signal from written text (text2speech). The only major category not dealing with text generation is `text representations' which deals with either word embeddings \cite{Mikolov:2013} or sentence embeddings \cite{Devlin:2018}, that is, vector representations of either words or sentences that summarize their semantic and syntactic properties. One speculation is that this latter field, text representations, will now be gradually declining in importance, after having dominated most of the 2010s. The more high-level topic text generation appears to replace it in its leadership role. 

\begin{figure}
    \centering
    \includegraphics[scale=.7]{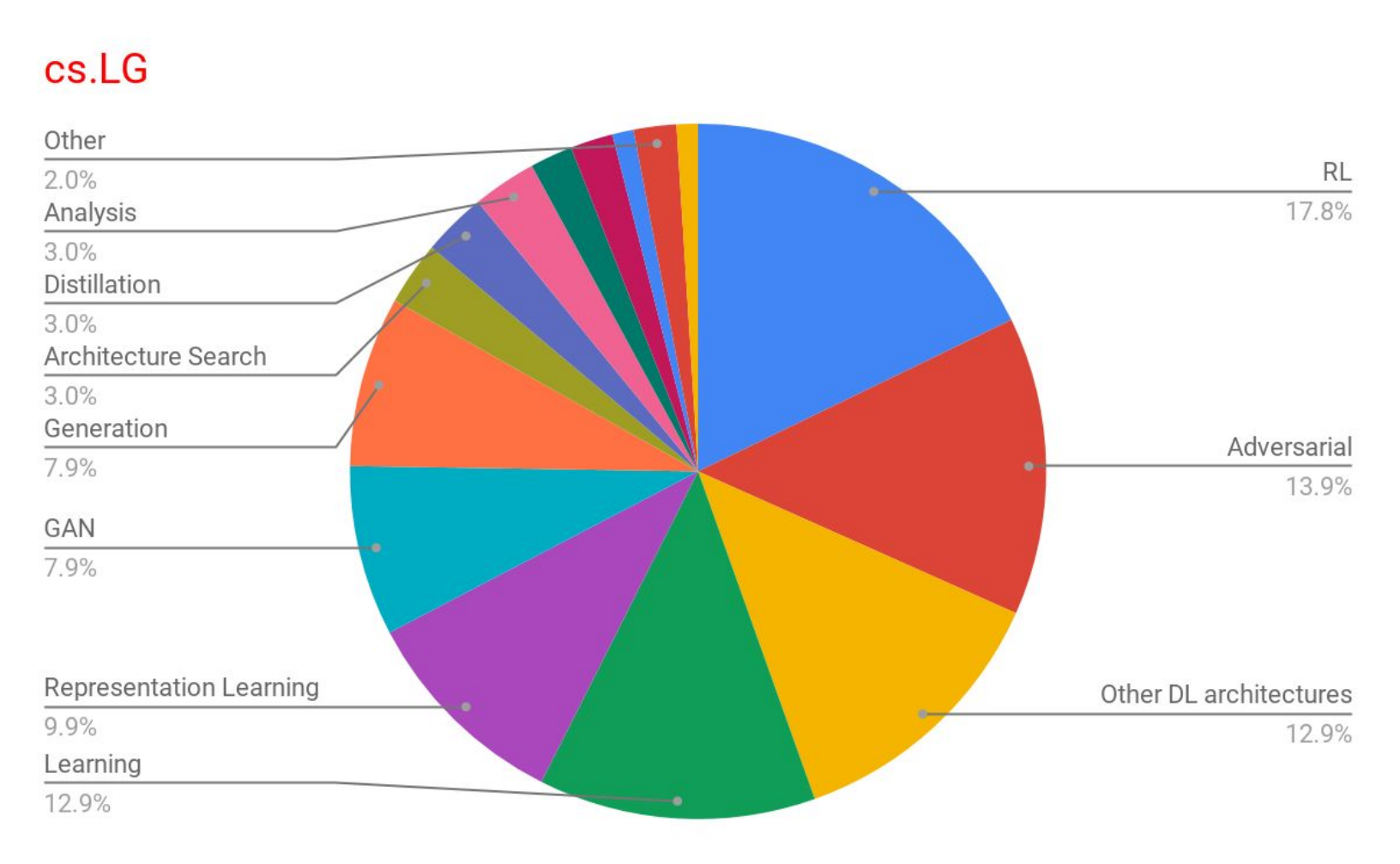}
    \caption{\texttt{method} distribution in top-100 list in cs.LG.}
    \label{figure:csLG}
\end{figure}

Figure \ref{figure:csLG} shows the \texttt{method} distribution in the top-100 list of cs.LG. The most important methods here are Reinforcement Learning (RL) and Adversarial techniques. The latter either deal with adversarial attacks on neural networks---i.e., fooling these networks into misclassification or shielding from such fooling---or with adversarial training, which aims to make machine learning systems more robust. Generation is likewise prominent, through Generative Adversarial Networks (GANs), which have been used to generate, e.g., artificial images with impressive quality. Indeed, the leading paper in the top-100 list uses GANs to generate `fake celebrities' \cite{Karras:2018}. Another interesting observation is that Reinforcement Learning is much less prominent in cs.CL (only 5\% of all papers indicate to use this method in their abstracts). It remains open whether NLP papers will catch up with the trends in general machine learning or whether reinforcement learning is generally less suited for natural language processing problems. 

\begin{figure}
    \centering
    \includegraphics[scale=0.7]{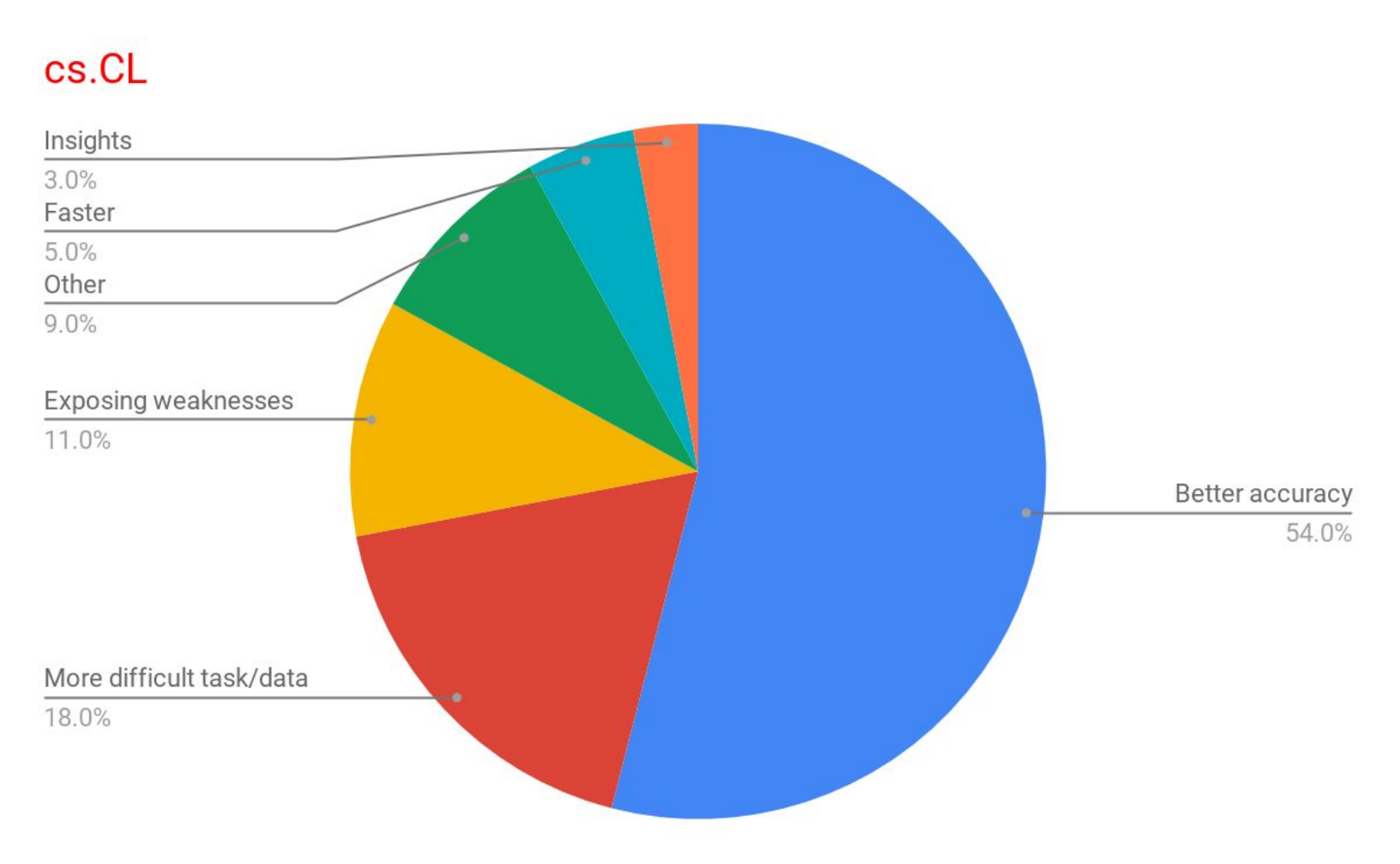}
    \caption{\texttt{goal} distribution in top-100 list in cs.CL.}
    \label{figure:CL_goals}
\end{figure}

Finally, Figure \ref{figure:CL_goals} shows the distribution of \texttt{goal}s in the top-100 list in cs.CL. About half of the papers present a better system that outperforms the previous state-of-the-art. Another major motivation in our cs.CL list is to create new and better datasets, which cover more realistic conditions, for example. Finally, exposing weaknesses either of NLP systems---such as their failure to adversarial attacks---or of researchers---such as drawing insufficiently supported conclusions regarding superiority of systems---are further main goals stated in the paper abstracts. 

\section{Related work}

Analyzing and predicting research trends is of interest to a variety of stakeholders. For example, Clarivate Analytics publishes a yearly ``Research fronts'' report discussing leading scientific topics.\footnote{https://clarivate.com/essays/research-fronts/} 
Their analysis also uses citation count statistics, even though these are not normalized as in the z-scores that we use.
A key difference to such studies is the database we use, namely, Arxiv. Arxiv is arguably a lead market place for the dissemination of machine learning research connected to deep learning (as a main driver of artificial intelligence research). 

There are other methods discussed in the literature for trend detection besides those based on citations. The authors of \cite{Mukherjee:2017} use the age distribution of the references of a paper to identify prospectively highly cited papers. The authors of \cite{prabhakaran2016predicting} use the rhetorical status of topics in paper abstracts to identify topics that will increase or decline over time. Further studies combine citation network analysis with temporal topic distributions obtained from LDA \cite{He:2009}, model the success of a paper by the centrality of its authors in co-authorship networks \cite{Sarigol:2014}, or track trends by analyzing keyword networks \cite{Duvvuru:2012}. 
\section{Concluding remarks}

We have developed a system to rank Arxiv papers according to their z-score, in order to detect short- and mid-term research trends. Our manual evaluation and clustering of ranked papers indicated some potentially interesting paradigm shifts in machine learning and NLP in the upcoming years. It appears that the fields will be more heavily concerned with fooling and attacking deep learning systems, as well as defending against such attacks, with data generation (instead of classification, as in traditional machine learning and NLP research), and with (deep) reinforcement learning, which can be seen as another endeavor towards proper artificial intelligence where artificial agents learn in a way similar to humans, 
performing actions and receiving rewards for them.\footnote{This analysis assumes that the normalized citation counts we use indeed help us to identify future citation behavior.} 

Further automatization of our methodology is necessary: we want to assign topics to ranked Arxiv papers automatically rather than manually. We will use neural text embeddings \cite{Devlin:2018}, keyword analysis \cite{Florescu:2017} and ontologies  \cite{Salatino:2018} for this. In the same vein or alternatively, clustering of Arxiv papers must be done automatically. As a secondary signal next to the z-score ranks, we want to predict citation counts from a model that incorporates the text and the meta-data (such as authors and publication venue) of a paper. Finally, we want to conduct a diachronic study: how does the ranking of high z-score papers/topics change over time? This may be considered as an additional layer of estimating and quantifying trends and their shifts. 

A final note is on our data. As with each dataset, Arxiv has its particular biases. One such bias is its connection to deep learning (due to deep learning's fast pace, which makes it an ideal candidate for a pre-print server). Indeed, almost all papers in our top-100 lists are about deep learning. As such, the topic distributions obtained from Arxiv may look considerably different from those derived from standard machine learning or NLP conferences---such conferences have their own biases, namely, to achieve a certain balancedness and democracy in choosing topics, which may in turn not truely reflect where the action is. 

Our tool plus the annotated data and the ranked Arxiv lists are available at \url{https://github.com/UKPLab/refresh2018-predicting-trends-from-arxiv}

%
%
%
\bibliographystyle{splncs04}
\bibliography{sample.bib}
%

%
%
%
\end{document}